\documentclass{sig-alternate-2013}

\usepackage{graphicx}
\usepackage[caption=false]{subfig}
\usepackage{color}
\usepackage{comment}
\usepackage{xspace}
\usepackage[linesnumbered, ruled]{algorithm2e}\DontPrintSemicolon
\usepackage{booktabs}
\usepackage{mdwlist}
\usepackage{url}
\usepackage{multirow}
\usepackage{minibox}
\usepackage{paralist}

\usepackage{amsfonts}
\usepackage{booktabs}
\usepackage{siunitx}

\renewcommand{\vec}[1]{\mathbf{#1}} %% vector
\newcommand{\mat}[1]{\mathbf{#1}} %% matrix

\newcommand{\hide}[1]{}

\newfont{\mycrnotice}{ptmr8t at 7pt}
\newfont{\myconfname}{ptmri8t at 7pt}
\let\crnotice\mycrnotice%
\let\confname\myconfname%

\begin{document}

%\conferenceinfo{KDD}{'15 Sydney, Australia}

%\title{Semi-supervised Learning with Deep Generative Models for Machine Failure Prediction}
\title{Semi-supervised Learning with Deep Generative Models for Asset Failure Prediction}

\numberofauthors{7}
\author{
%\alignauthor
Andre S. Yoon, Taehoon Lee, Yongsub Lim, Deokwoo Jung \\
Philgyun Kang, Dongwon Kim, Keuntae Park, Yongjin Choi \\
\affaddr{Big Data Tech. Lab}\\
\affaddr{SK Telecom}\\
%Andre S. Yoon\\
%\alignauthor
%Taehoon Lee\\
%\alignauthor
%Yongsub Lim\\
%       \affaddr{Department of Computer Science}\\
%       \affaddr{School of Computing}\\
%       \affaddr{KAIST}\\
%       \email{yongsub@kaist.ac.kr}
%\alignauthor
%U Kang\\
%%       \affaddr{Department of Computer Science}\\
%       \affaddr{School of Computing}\\
%       \affaddr{KAIST}\\
%       \email{ukang@cs.kaist.ac.kr}
}

\maketitle
\begin{abstract}

%% Short-version
This work presents a novel semi-supervised learning approach for data-driven modeling of asset failures when health status is only partially known in historical data. We combine a generative model parameterized by deep neural networks with non-linear embedding technique. It allows us to build prognostic models with the limited amount of health status information for the precise prediction of future asset reliability. The proposed method is evaluated on a publicly available dataset for remaining useful life (RUL) estimation, which shows significant improvement even when a fraction of the data with known health status is as sparse as 1\% of the total. Our study suggests that the non-linear embedding based on a deep generative model can efficiently regularize a complex model with deep architectures while achieving high prediction accuracy that is far less sensitive to the availability of health status information.

\end{abstract}

% A category with the (minimum) three required fields
\vspace{-1mm}
%\category{G.2.2}{Graph Theory}{Graph algorithms}
%A category including the fourth, optional field follows...
%\category{H.2.8}{Database Applications}{Data mining}

\vspace{-1mm}
%\terms{Design, Experimentation, Algorithms}
%\terms{Design, Experimentation, Algorithms}

\vspace{-1mm}
%\keywords{Local triangle counting; graph stream mining; edge sampling; anomaly detection}

\section{Introduction}
% Look for SUF, REF,

% Below paragraph should be polished/corrected with actual references!

% https://www.hindawi.com/journals/mpe/2015/793161/  <=== Very useful
% http://www.sciencedirect.com/science/article/pii/S0377221710007903
% http://www.sciencedirect.com/science/article/pii/S0888327010004218

% https://ti.arc.nasa.gov/tech/dash/pcoe/data-driven-prognostics/
% https://tel.archives-ouvertes.fr/tel-01126861/document
% https://www.phmsociety.org/sites/phmsociety.org/files/Saxena_Prognostics_TutorialPHM10.pdf

%% 용어 통일할것! Machine, asset, reliability, RUL, health index?
%% 어체 통일할거: Passive & active

%% Data-driven approaches in RUL estimations
%% http://www.sciencedirect.com/science/article/pii/S0377221710007903

With increasingly more sensor data available either transient in stream or permanent in database, numerous physical assets in different industrial sectors such as manufacturing, transportation, and energy are subjected to asset condition and health monitoring~\cite{kwon2016iot}. When sufficient information present in the data, referred to as condition monitoring (CM) data, the asset health status and its evolution can be modeled to predict a specific point in time after which an asset will no longer meet its normal operational requirements, i.e., asset failure~\cite{kwok15}. The concept of RUL, also referred to as an estimated time to failure, is widely used in the future reliability prediction with an important industrial application known as predictive maintenance (PdM)~\cite{mobley2002introduction}. The PdM attempts to reduce costly unscheduled maintenance or maintenance that is excessively scheduled.

Existing approaches for predictive maintenance belong two categories: physics-driven (deduction from physical equations underlying mechanical operations) and data-driven (induction from measurements under actual operations). In this paper, we narrow down our scope to only focus on data-driven methods. Gaussian and Markov processes have produced promising results~\cite{kwok15, si2011remaining, sikorska2011} in the field of predictive maintenance since 1990's, but they often suffer from practical limitations such as scalability (\textit{i.e.}, a basic form of Gaussian process requires $O(n^3)$). On small datasets, they are very useful because strong priors can cover lack of data with reasonable computations. However, as data volumes are exploding, more complicated functions are required to model massive, high-dimensional operational measurements which are hard to be characterized by well-known priors and kernel functions. 
%high-dimensional and massive operational measurements which are hard to be characterized by well-known priors and kernel functions. 
%researchers need to more complicated functions underlying high-dimensional and massive operational measurements which are hard to be characterized by well-known priors and kernel functions. 
%The advent of deep learning enables us to capture the complicated functions; for example,
The advent of deep learning has enabled us to capture the complicated functions; for example, \cite{zhao2016deep} exhibited performance improvements with deep learning-based prognostic models to predict the RUL of given assets.

%[Little more about the DL-based approaches: e.g., no feature extractions]

While many deep learning-based approaches rely on the abundance of historical data with the availability of corresponding health indicators either from the CM data itself or from extra measurements or modelings, obtaining exact health status of a given asset in reality is highly expensive or impossible in some cases. It is because labeling requires expertise or special devices to produce it, and generally highly time consuming. % compared to the operational time of an asset.
Above all, failures are rare and not easily replicated, especially for those that are mission-critical. This is also closely related to the problem of intermittent failures also known as No Fault Found (NFF) problem~\cite{kwok15}. For example, in the case of mission-critical assets such as an aircraft engine or equipment in nuclear power plant, even a single failure is catastrophic and therefore, regular maintenance or replacement is carried out regardless of the true condition of an asset. Or in the case of fresh start of newly installed assets, one may have to wait long to collect any run-to-failure data to precisely model the degradation and consequently failure.

In the lack of sufficient health status information, which we call a label or a target variable, $y$, data-driven modeling of degradation becomes immensely difficult in many cases. It is often the case where there is abundant data with relatively few labels since unlabeled data is relatively cheap. We recognize while the complexity of various data-driven models increases, especially with the latest deep neural net architectures, there has been relatively little attention paid to the problem of insufficient labels in data-driven prognostics. In the field of machine learning, researchers have long recognized such difficulty and explored various techniques to make learning better generalized in the lack of sufficient labels~\cite{ssl2010}.

Semi-supervised learning (SSL) is a class of techniques that falls between unsupervised learning and supervised learning (SL)~\cite{ssl2010}. While supervised learning relies solely on labeled data, SSL makes use of unlabeled data as well to improve learning accuracy. Thus, SSL is resorted to deal with a situation where relatively few labeled data are available with large unlabeled data.
There exist many different approaches in SSL, which include heuristic approaches such as self-learning and approaches based on generative models, and low-density separation~\cite{ssl2010}. Among different approaches, generative models parameterized by deep neural networks referred to as deep generative models (DGM) have recently achieved state-of-the-art performance in various unsupervised and semi-supervised learning tasks~\cite{dgm2014, aux2016, ladder2015}. Two most popular approaches of DGM rely on variational inference~\cite{vae2013} and generative adversarial network (GAN)~\cite{gan2014}. The latter being probably the most popular in deep learning research nowadays is grounded in game theory.

In this paper, we propose a semi-supervised learning technique to improve data-driven asset failure prediction when labeled data is highly limited compared to the unlabeled. For example, we consider the case where historical CM data are abundant from a fleet of assets but it is only partially labeled with health status.
%For example, we consider the case where historical CM data from normal ope with smaller historical CM data from failed assets. where historical CM data are abundant from a fleet of assets but it is only partially labeled.
In such a case, we observe the prediction accuracy becomes highly degraded even with sophisticated deep learning models based on the Recurrent Neural Networks (RNN) architectures such as Gated Recurrent Unit (GRU) or Long Short-Term Memory (LSTM). Our proposed method uses a non-linear embedding, which is obtained from one of the DGMs known as variational autoencoder (VAE). The variational autoencoder makes use of all the available data, both labeled and unlabeled. A part of the VAE is taken to form an embedding network, which is used to define a latent space where a model for reliability prediction is trained.

In the remainder of this paper, we first give the background of this work in Section 2 by discussing deep neural networks and a variational autoencoder as a choice of deep generative model. We describe our proposed methods in Section 3. In Section 4, we further the details of our methods with the experimental evaluations. Lastly, we discuss our findings in Section 5.

\label{intro}

%\section{Motivation and Approach Overview}
%\section{Approach Overview}
\section{Background}
\label{sec:overview}

\subsection{Problem Statement}
\label{sec:pre}
In our problem, we have multivariate data for sensor measurement and their corresponding RUL.
Let $x^{(i,j)}$  denote $i_{th}$ measurement value of sensor $j$.
Furthermore, let us $\vec{x^{(i)}}$ denote a vector of multivariate sensor measurement
such that $\vec{x^{(i)}}=[x^{(i,1)},\cdots, x^{(i,m)}]$ where $m$ is the number of sensors.
Formally, we describe a sensor measurement matrix by denoting
$\mat{x}=\{\mathbf{x^{(1)}},\cdots,  \mathbf{x^{(n)}} \}$  where $\mathbf{x^{(i)}} \in \mathbb{R}^m$
and $n$ is the number of observed measurements.
We also denote the corresponding set of RUL by
 $\mathbf{y}= \{y^{(1)},\cdots,  y^{(n)}  \} $ where $y^{(i)} \in \mathbb{R}^+$.
More compactly, we define the data set $\mathcal{D}$ by $\mathcal{D} =\{(\mathbf{x}^{(i)},y^{(i)})\}_{i=1,\cdots,n}$.

In general, the data-driven prognostics approach is to learn the best predictor of RUL (or equivalent health indicator used to predict RUL) from the previously observed data set $\mathcal{D}^{T}$, i.e.training data set. The challenging problem is to find an appropriate transformation of the raw measurement $\mathbf{X}$ to efficiently learn a reliable RUL predictor. Such problem can be described by finding a mapping function $F$ such that $F :\mathbf{x} \mapsto  \mathbf{z}$ where a latent variable $\mathbf{z} \in \mathbb{R}^k$ and  $m<k$.
%Then the optimal predictor can be defined by a function of $\mathbf{z}$ with parameter $\mathbf{\theta}$ denoted by
%$r_{\mathbf{\theta}}( \mathbf{z})$   such that $r_{\mathbf{\theta}}( \mathbf{z})= \underset{y}{max}~ p( y~|~\mathbf{z},\mathbf{\theta}) $.
Then the optimal predictor can be defined by a function of $\mathbf{z}$ with parameter $\mathbf{\gamma}$ denoted by
$f_{\mathbf{\gamma}}( \mathbf{z})$   such that $f_{\mathbf{\gamma}}( \mathbf{z})= \underset{y}{argmax}~ p( y~|~\mathbf{z},\mathbf{\gamma}) $.

In traditional machine learning approaches, it first finds appropriate latent variable $\mathbf{z}$ by well known linear transformation methods (e.g. Principal Component Analysis ) often with heuristics.
Then it estimates the optimal parameter $\mathbf{\gamma}$ by training $f_{\mathbf{\gamma}}( \mathbf{z})$ with $\mathcal{D}^{T}$ given a certain simplified assumption for the distribution
$p( y~|~\mathbf{z},\mathbf{\gamma})$(e.g. multivariate normal distribution).
Such conventional approaches inherently has its own limitations in providing a general solution for RUL prediction. First, it is often that the mapping function $F$ for latent variable $\mathbf{z}$ is non-linear and its specific form is largely unknown. Second, any strong assumption on distribution
$p( y~|~\mathbf{z},\mathbf{\gamma})$ often poorly captures complex cross dependency over time and sensors. Meanwhile, deep neural network can provide a generic solution to jointly learn the non-linear mapping function $F$ and the predictor $f_{\mathbf{\gamma}}$ from training data set $\mathcal{D}^{T}$ using directed graph network with various levels of abstraction. We drop the superscript $T$ for the rest of the paper.

In our problem setting, we assume that only a small fraction of the samples has label information. Let $\mathcal{D}^{T}_{L}$ and  $\mathcal{D}^{T}_{U}$ denote the part of the data set with labels and the part without the labels, respectively. The primary goal in this paper is to develop deep learning-based approach to predict RUL from $\mathcal{D}^{T}_{L}$ where $|\mathcal{D}^{T}_{L}| \ll |\mathcal{D}^{T}_{U}|$, which is a class of semi-supervised learning. In the semi-supervised learning setting, we assume that $\mathcal{D}^{T}_{L}$ comes from exactly the same distribution of $\mathcal{D}^{T}_{U}$.
Our semi-supervised approach fully exploits $\mathcal{D}^{T}_{U}$ to learn the non-linear mapping function $f$ to get a robust feature representation of $\mathbf{x}$  for stable RUL predictor $f_{\mathbf{\gamma}}$  given $\mathcal{D}^{T}_{L}$.

%\begin{figure}
%\begin{center}
%	\includegraphics[width=0.99\linewidth]{FIG/sensor_as_cycles.pdf} \\ \vspace{-3mm}
%	\caption{Pearson correlation coefficient $\rho$}
%	\label{fig:metrics}
%\end{center}
%\end{figure}

\subsection{Neural Network Overview}\label{dnn_overview}
%% Citation!!!
\begin{comment}
% This part is for semi-supervised learning
Semi-supervised learning (SSL) aims to improve generalization on supervised tasks using unlabled data. In the semi-supervised learning framework, a given set of $N = L+U$ ($U >> L$ usually)  data can be divided into two parts: historical data $X_{l} = (x_{1},..,x_{l})$ for which corresponding labels $Y_{l} = (y_{1},..,y_{l})$ exist and $Xu = (x_{l+1},..,x_{l+u})$, for which corresponding labels do not exist. The goal of SSL is to improve prediction accuracy of a function learned from $(X_{l}, Y_{l})$, i.e., by taking into account the unlabeled data $X_{u}$. There are several assumptions that a number of SSL theories are based on. These are are smoothness, cluster, and manifold assumptions, each of which is related to the structure of the underlying distribution. While these assumptions are of great theoretical interest and may provide intuitions into SSL, we will not go into further details as our intention is not to uncover theoretical maximum performance gain that SSL approach can bring per se.
\end{comment}

In this section we briefly discuss the definition and the architectures of basic deep neural network and its variants.

\subsubsection{feed-forward neural network} \label{ffnw}
%In neural network, a neuron is a basic building block to form a larger complex network of information flows. A neuron computes a weighted sum of children neurons' outputs and selectively feed it to its parent neurons. The computation is done by two activation functions ; linear pre-activation function for the weighted sum and non-linear activation function for the selective feed.
%A neural network is simply a directed graph of neurons that allows information to propagate over network with a great degree of freedom in its direction and magnitude. A feed-forward neural network is the basic form of neural network that consists of a number of layered neurons without cycle in the network. When the number of layers goes large, it is called deep neural network.

We can formally describe a feed-forward neural network as follows.
Let  $f_{l,i}$ and $g_{l,i}$ denote a linear pre-activation and non-linear activation function of $i$th neuron at layer $l$, respectively where  $f_{l,i},g_{l,i} \in \mathbb{R}^k$ for $i=1,\cdots, n_l$,  $n_l$ is  the number of neurons at layer $l$, and $k$ is the dimension of a neuron's outputs.
We can conveniently assume that the output dimension $k$ is the dimension of a single sensor measurement.
In our problem setting, $k=1$ since each sensor measurement is a scalar value.
Then layer $l$ has two function vectors ; $f_l=[f_{l,1}\cdots f_{l,n_l}]^T $ and  $g_l=[g_{l,1}\cdots g_{l,n_l}]^T$.

Let $\mathbf{x}_l=[\mathbf{x}_{l,i}]_{i=1,\cdots,n_l}$ denote activation output at layer $l$ where  $\mathbf{x}_{l,i}=g_{l,i}\circ f_{l,i}(\mathbf{x}_{l-1})$.
The pre-activation of layer $l$ is computed by linear transformation
\begin{equation}\label{eqn:actfunc}
  \mathbf{f}_l(\mathbf{x}_{l-1})=\mathbf{W}_l \mathbf{x}_{l-1}+\mathbf{b}_l
\end{equation}
where  $\mathbf{W}_l \in \mathbb{R}^{ n_l \times n_{l-1}}$ and  $\mathbf{b}_l \in \mathbb{R}^{n_{l-1}}$.

For the non-linear activation, we use a popular rectified linear unit (ReLU)~\cite{NairH10} where
$g_i(f_i) =\max \{0, f_i\}$ for  $f_i \in \mathbb{R}$.
%The computation $\mathbf{f}_l(\mathbf{x}_{l-1})$ for $i=1\cdots L$ is done by forward propagation
The layer $l$'s parameter, $\theta_l=(\mathbf{W}_l,\mathbf{b}_l)$ is estimated by the\emph{ backpropagation algorithm} \cite{LeCun98} given $\mathcal{D}^{T}$.
Finally, feed-forward neural network of $L$ layers is defined as following :
\begin{equation}\label{eqn:ffnn}
  F(\mathbf{x};\theta)=f_{out}\circ h_L(\mathbf{x})
\end{equation}
%where  $h_l= (g_l \circ f_l)\circ h_{l-1}$ and $\theta=(\theta_l,\cdots,\theta_L) $.
where  $h_l= (g_l \circ f_l)\circ h_{l-1}$ and $\theta=(\theta_{0},\cdots,\theta_L) $.

\begin{figure*}[t!]
\begin{center}
	\includegraphics[width=0.95\linewidth]{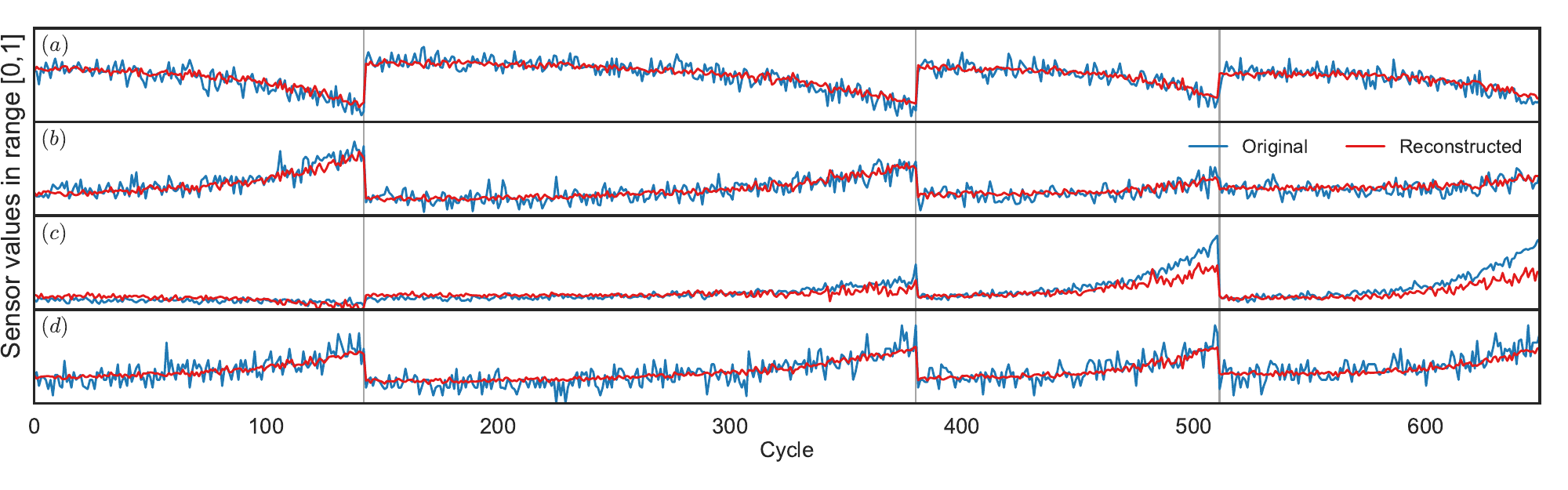} \\ \vspace{-3mm}
	\caption{Normalized sensor values as a function of cycles for four selected sensors $(a)$-$(d)$ from four independent engine: original (blue) and reconstructed (red). The cycles are cumulative. The gray lines drawn to differentiate the units.}
	\label{fig:cycles}
\end{center}
\end{figure*}

\subsubsection{Recurrent Neural Networks}
%% Citation!!!

The feed-forward neural network we described in previous section does not allow
temporal dependency of samples to be embedded in its model.
In our problem, the current sample are mostly likely to be dependent on its previous samples.
Hence, we need a more generalized model capable of elegantly incorporating
temporal dependency in it. RNN is such kind of a model that can capture into model the influence of an arbitrary number of previous samples on the current~\cite{dlbook2016}.
The basic RNN structure consists of a cell with a cyclic loop whose internal state evolves over time by the current sample input and its previous state output. The output of RNN is simply a series of cell states.
When we represents the cell state at each time as a neuron, its resultant structure is a non-cyclic horizontally connected neural network where the current neuron's output feeds into the next neuron's input.

Using the horizontal representation of cell states we can formally describe RNN with a simple modification of (\ref{eqn:actfunc}) and (\ref{eqn:ffnn}).
Let us assume that we use the previous  $T$ samples for RNN input.
Then the RNN input at time $t$ is the array of multivariate sensor measurements from $t$ to $t_T$ denoted by
$\mat{x}^{t}_{t-T}=[\vec{x}^{t},\cdots,\vec{x}^{t-T}]$.
Let $\vec{s}_t$ denote a cell state at time $t$ such that $\vec{s}_{t} \in \mathbb{R}^{k}$ where $k$ is a dimension of the state. Then pre-activation function of a neuron is defined by a linear mapping of the cell's previous state $\vec{s}_{t-1}$  and the current input  $\vec{x}_{t} \in \mathbb{R}^{m}$  at time $t$  as shown in  (\ref{eqn:rnn_actfunc}).
\begin{equation}\label{eqn:rnn_actfunc}
  f(\vec{s}_{t-1},\vec{x}_{t})=\mat{U}\vec{s}_{t-1}+\mat{W}\vec{x}_{t}+\vec{b}
\end{equation}
where $\vec{b} \in \mathbb{R}^{m}$, $\mat{U} \in \mathbb{R}^{ k \times k}$, and  $\mat{W} \in \mathbb{R}^{ k \times m}$.

The cell state at time $t$ is updated by a non-linear activation function $g$ on the $f$ output
such that $\vec{s}_{t}=g\circ f(\vec{s}_{t-1},\vec{x}_{t})=h(\vec{s}_{t-1},\vec{x}_{t})$.
Finally, RNN for $T_{th}$ order time dependency is compactly defined as following :
\begin{equation}\label{eqn:rnn_model}
  F(\mat{x}^{t}_{t-T};\mat{\theta})=[ h(\vec{s}_{t'-1},\vec{x}_{t'}) ]_{t'=t,\cdots,t-T}
\end{equation}
where  $h = (g \circ f)$ and
$\mat{\theta} =(\mat{U},\mat{W},\mat{b})$.
Note that all neural cells share parameter $\mat{\theta}$ over time.

The RNN parameter $\mat{\theta}$ can be trained using backpropagation through time (BPTT) algorithm~\cite{Werbos90}.
However, RNNs with the BPTT suffer from the \emph{vanishing} and \emph{exploding} gradient problem~\cite{Bengio94} because a recurrent network grows as deep as sequence length. One of popular solutions is to exploit Long Short-Term Memory (LSTM). An LSTM~\cite{Hochreiter97} is a second-order recurrent network that has three additional gate layers called \textit{input}, \textit{forget}, and \textit{update} gates. The gate parameters helps a recurrent network have good information flow over time by alleviating the vanishing gradient. Thus, a recurrent network \textit{with} gates can work better than one \textit{without} gates as more long-term dependencies exist.

GRU~\cite{cho14} is a simpler variant of LSTM which combines \emph{forget input} gate and \emph{read input} gate into \emph{overwrite} gate by setting an input gate to the inverse of forget gate, ,i.e $i_t=1-f_t$. It is shown that the performance of GRU is on par with LSTM  but more computationally efficient~\cite{Rafal15}. In our study, we find there is no meaningful difference in prediction accuracy between the two variants.
%We use GRU instead of LSTM for more efficient implementation of RNN.

\begin{figure}[h]
\begin{center}
	\includegraphics[width=0.9\linewidth]{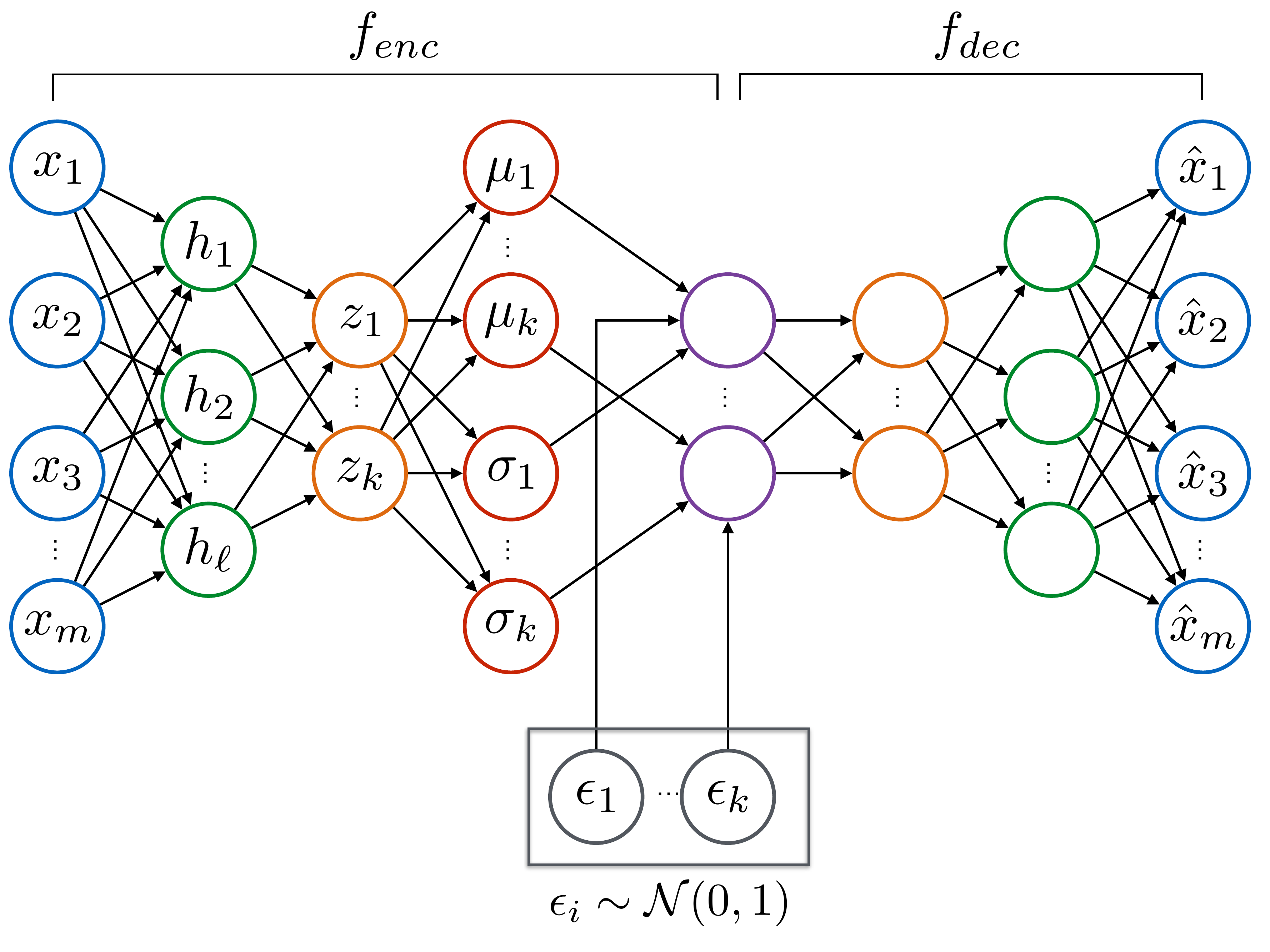}  \\ \vspace{-1mm}
	\caption{Overall architecture of VAE with an input dimension of $m$. $\epsilon \sim \mathcal{N}(0,1)$ indicates the reparameterization trick is used in the training of VAE.}
	\label{fig:vae}
\end{center}
\end{figure}

\begin{figure}[h]
\begin{center}
	\subfloat[\label{mae} Embedding Network]{\includegraphics[width=0.48\linewidth]{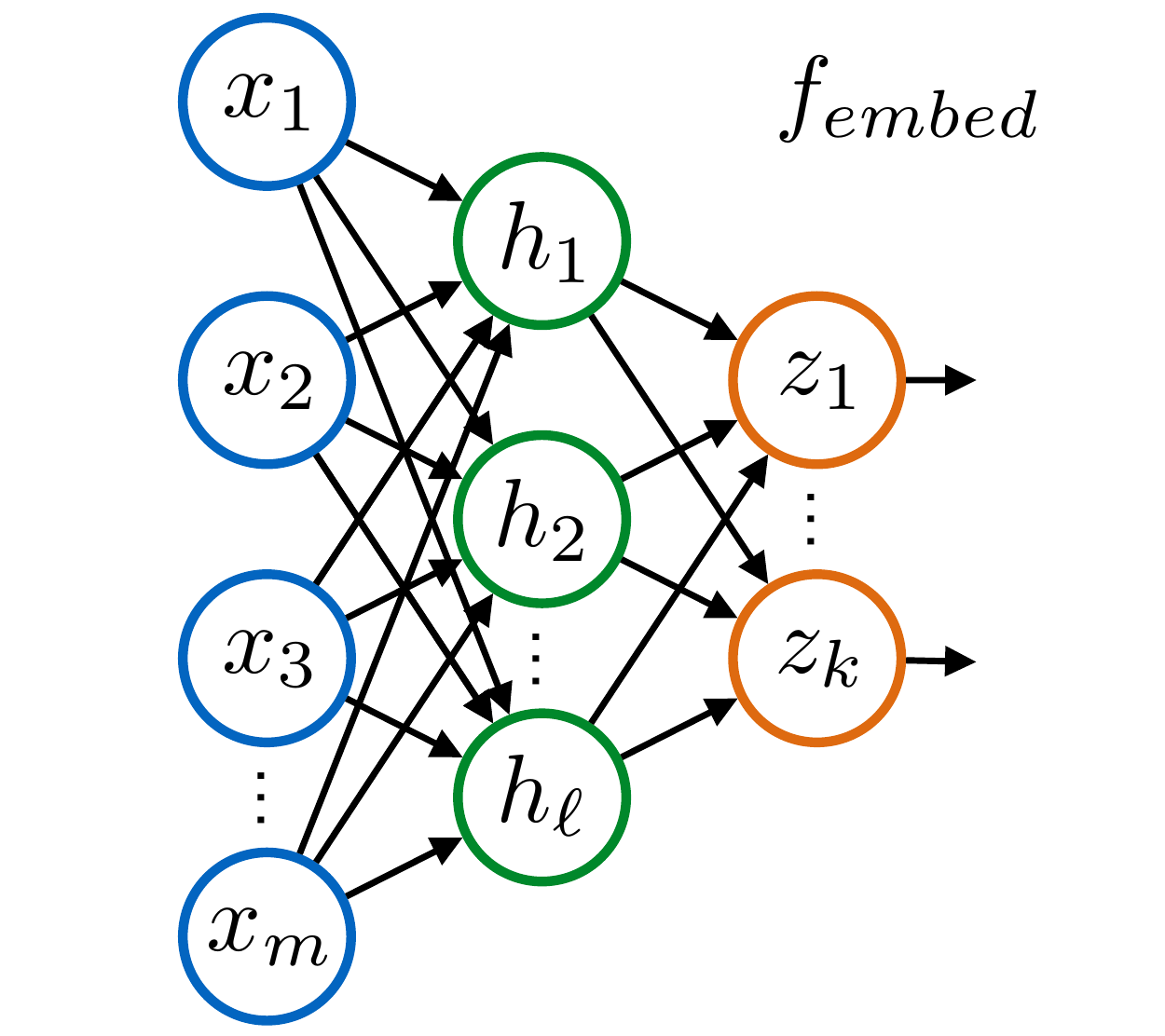}}
	\subfloat[\label{mse} Reliability Model]{\includegraphics[width=0.48\linewidth]{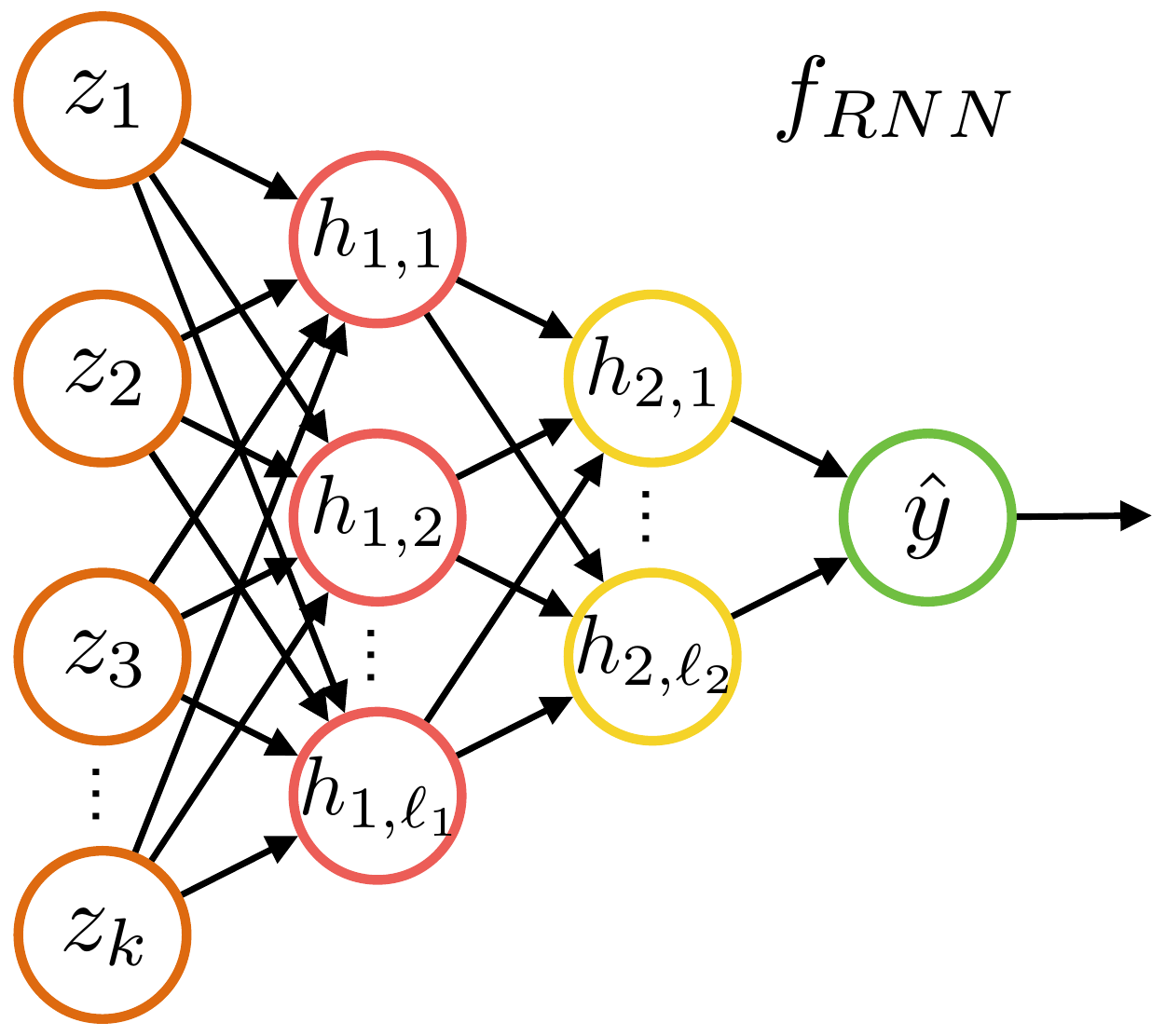}} \\ \vspace{-1mm}
	\caption{Embedding followed by supervised learning in latent space. (a) embedding network is taken from the encoder of the trained VAE in Fig.~\ref{fig:vae}. (b) reliability model is built upon the latent space defined by the embedding network. While RNN architecture is used, the recurrence relation is not shown for simplicity.}
	\label{fig:embedding}
\end{center}
\end{figure}

%\subsubsection{Variational Autoencoder (VAE)}
\subsection{Variational Autoencoder (VAE)}\label{vae}
% http://www.cs.toronto.edu/~slwang/generative_model.pdf
% dlbook2016, hinton2006
% vae2013, vae2013v2

% 0. loss function (objective)
% 1. SGVB algorithm
% 2.

An autoencoder is a neural net with an encoder-decoder architecture that is trained to reconstruct its own input. Given an input $\vec{x}$, the autoencoder, $f_{enc}(\vec{x})$, encodes and then the decoder, $f_{dec}(\vec{z})$, decodes the encoded input (a \emph{representation}) back to reconstruct the input as closely as possible by minimizing a loss function of the form~\cite{dlbook2016}:
\begin{equation}\label{eqn:aeloss}
	\mathcal{L}(\vec{x}, \vec{x}') = || \vec{x} - f_{dec}(f_{enc}(\vec{x})) ||
\end{equation}
where $f_{enc}$ and $f_{dec}$ are symmetric in topology.
%are the feed-forward neural networks as described in Section~\ref{ffnw}.
The autoencoder is trained using backpropagation algorithm in the same manner as in the feed-forward neural network in Section~\ref{ffnw}. However, when the architecture becomes deep with many hidden layers, pretraining is necessary to set the initial weights of the autoencoder close to the final ones~\cite{hinton2006}. The VAE inherits autoencoder's encoder-decoder architecture, but it is different in that the latent representation $\vec{z}$ of given data $\vec{x}$ are replaced with stochastic variables. The encoder and the decoder of VAE are probabilistic and given by $q_{\phi}(\vec{z}|\vec{x})$, an approximate posterior, and $p_{\vec{\theta}}(\vec{x}|\vec{z})$, likelihood of the data $\vec{x}$ given the latent variable $\vec{z}$, respectively. The approximate posterior is parameterized by a neural network and the likelihood is given by a multivariate Gaussian whose probabilities are obtained from $\vec{z}$.
%% Explain
%% 0. Approxmiate posterior
%% 1. likelihood
%e omit the index $i$ in the notation for the sake of simplicity.
%%%%%%%%%%%%%%%%%%%%%%%%%%%%%%%%%%%%%%%%
%% More on parameterziation of q and p parameterizations with DNN
%%%%%%%%%%%%%%%%%%%%%%%%%%%%%%%%%%%%%%%%

The objective function of VAE is the variational lower bound to the marginal likelihood of the data, i.e., $\log p_{\theta}(\vec{x}) = \sum_{i=1}^{n}\log p_{\theta}(\vec{x^{(i)}})$. It can be rewritten for individual data points, $\vec{x}^{(i)}$ as~\cite{vae2013}:
\begin{equation}\label{eqn:likelihood}
\log p_{\theta}(\vec{x}^{(i)}) = D_{KL}(q_{\phi}(\vec{z}|\vec{x}^{(i)})||p_{\theta}(\vec{z}|\vec{x}^{(i)})) + \mathcal{L}(\theta,\phi;\vec{x}^{(i)})
\end{equation}
where $D_{KL}$ is Kullback-Leibeler (KL) divergence. Since $D_{KL} \geq 0$, (\ref{eqn:likelihood}) can be rewritten as:
\begin{equation}\label{eqn:lowerbound}
\log p_{\theta}(\vec{x}^{(i)})  \geq \mathcal{L}(\theta,\phi;\vec{x}^{(i)})
\end{equation}
The left-hand side of (\ref{eqn:lowerbound}), which is called \emph{variational lowerbound}, can further be rewritten using Bayes' rule and the definition of KL divergence as~\cite{vae2013}:
\begin{equation}\label{eqn:lb2}
-D_{KL}(q_{\phi}(\vec{z}|\vec{x}^{(i)})||p_{\theta}(\vec{z})) + E_{q_{\phi}(\vec{z}|\textbf{x}^{(i)})}\big[\log p_{\theta}(\vec{x}^{(i)}|\vec{z}) \big]
\end{equation}
which can be viewed as the sum of reconstruction error plus a term that acts as a regularization.

%The training of VAE is done by optimizing (\ref{eqn:lb2}).

%While backpropagation is used to train the VAE,

While obtaining the gradients of the encoder is relatively straightforward, that of the decoder is not. This is because, the reconstruction error term in Eq.~\ref{eqn:lb2} requires the Monte Carlo estimate of the expectation, which is not easily differentiable~\cite{vae2013}. This problem is solved by introducing a reparameterization of $z$ with a \emph{deterministic variable} such that $z = \mu + \sigma\epsilon$, with $\epsilon \sim \mathcal{N}(0,1)$, which is known as ``reparameterization trick''~\cite{vae2013}. With this trick, a differentiable estimator of the variational lower bound can be obtained and the backpropagation algorithm can be applied to train VAE. The overall architecture of VAE is shown in Fig.~\ref{fig:vae}.

\begin{figure}[h]
\begin{center}
	\includegraphics[height=0.9\linewidth]{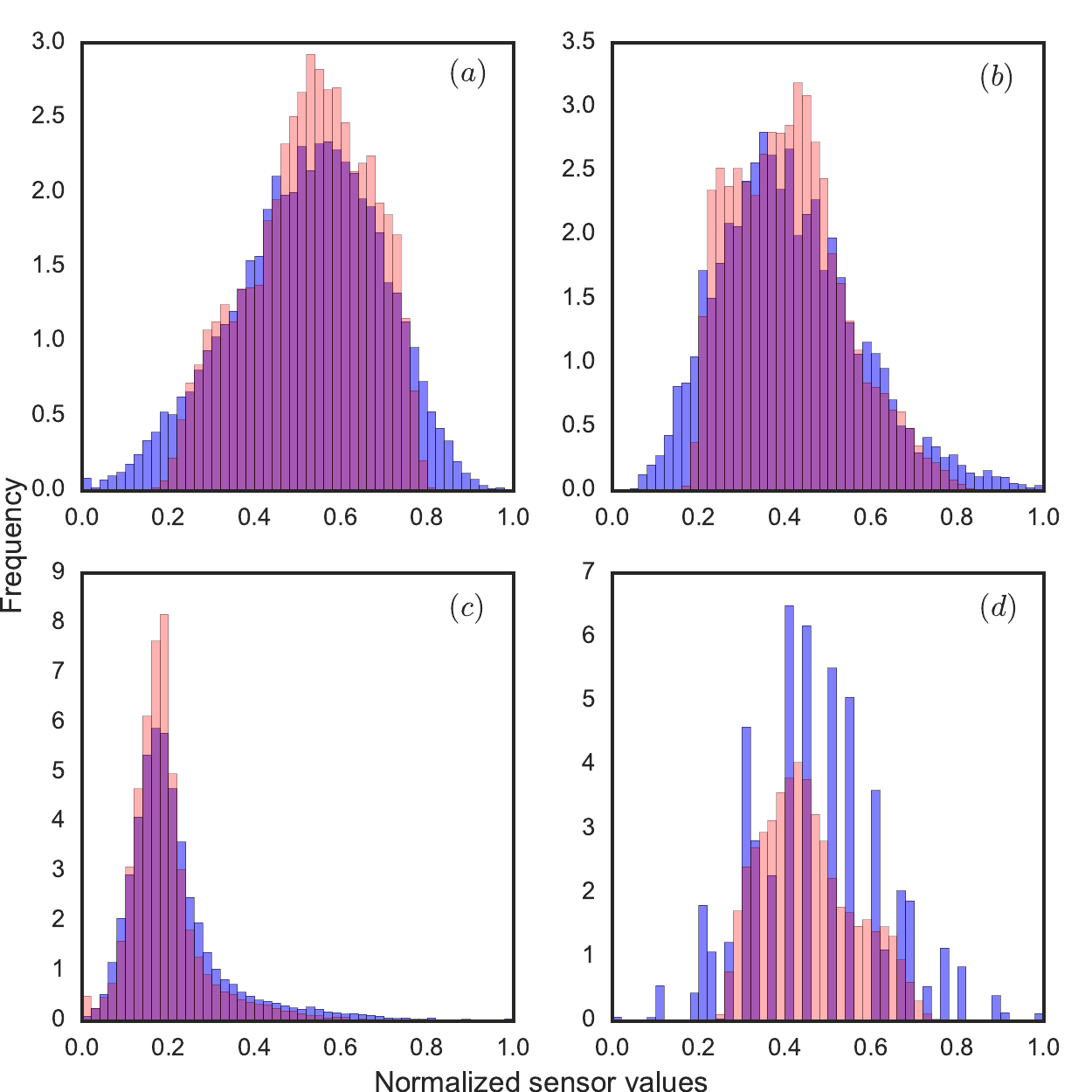} \\ \vspace{-1mm}
	\caption{Normalized frequency of the values of selected sensors in Fig~\ref{fig:cycles} $(a)$-$(b)$ from all the engines in the dataset: original (blue) and reconstructed (red).}
	\label{fig:freq}
\end{center}
\end{figure}

\section{Our Approach}
\subsection{Overview}\label{ourapproach}

%. Brief introduction to Non-linear embedding
%. What we choose, how we do? Algorithms + Self-learning
%  Architecture details (in th later section?)
%\subsubsection{Non-linear Embedding}
% based on embedding. What is embedding? what kinds?

%"The basic idea of our approach is to build a model for semi-supervised learning that exploits generative description of the data to improve upon the prediction accuracy that would be obtained using"

%The basic idea of our approach is to build a model for semi-supervised learning that exploits generative description of the data. The generative description of the data is obtained from the VAE and the  the model is trained in the latent space.
%There are different modes of embedding in deep neural net architectures as described in~\cite{weston2008}.

In this section, we describe our overall approach for semi-supervised learning for RUL prediction
using neural network framework and a deep generative model described in Section~\ref{dnn_overview} and~\ref{vae}. We provide a description of the implemented architectures for RUL prediction and several important architectural decisions on design parameters in Section~\ref{exp}.

%We describe how we designed the architecture for RUL prediction with emphasis on several important architectural decisions on design parameters in Section~\ref{exp}.

%In this section we describe how we designed neural net architectures for RUL prediction  with emphasis on several important architectural decisions on design parameters.

\subsubsection{Non-linear Embedding}
Embedding in semi-supervised learning is a procedure to map $\vec{x}$ in the original space into a lower dimensional space, $\vec{z}$, with the mapping obtained from unsupervised learning. The goal of embedding is to improve generalization of a supervised model. In our approach, we use the VAE to obtain the non-linear embedding. It is done by training the VAE as described in Section~\ref{vae} using $\vec{x}$ from both labeled $\mathcal{D}_{L}$ and unlabeled data $\mathcal{D}_{U}$ to learn a mapping function $f_{embed}$ for latent variable $\vec{z}$. The latent space is defined by the output layer chosen in the encoder of VAE as illustrated in Fig.~\ref{fig:embedding}(a). We choose $k^{th}$ layer of the encoder with $k$ less than $L_{encoder}$ (a total number of layers used in the encoder) to avoid using approximate samples from posterior distribution.

While for the parameterization of the approximate posterior in VAE, feed-forward neural network is conventionally chosen as originally suggested in~\cite{vae2013}. In our case, we choose the RNN-architecture for the parameterization to model the temporal dependency of data. Incorporating the RNN architecture into VAE to include the latent random variables into the hidden state of a RNN requires the estimation of timestep-wise variational lower bound, which is not trivial. We instead simply introduce a weight factor $\alpha$ such that the two terms in Eq.~\ref{eqn:lb2} is differently weighted while modeling the temporal dependence in data with the RNN architecture.
\begin{equation}\label{eqn:lb3}
-D_{KL}(q_{\phi}(\vec{z}|\vec{x}^{(i)})||p_{\theta}(\vec{z})) + \alpha E_{q_{\phi}(\vec{z}|\textbf{x}^{(i)})}\big[\log p_{\theta}(\vec{x}^{(i)}|\vec{z}) \big]
\end{equation}
The weight factor is tuned considering the overall reconstruction efficiency of VAE.

%As such, we incorporate sequentially modeling in VAE by introducing a weight factor such that the two terms in Eq.~\ref{eqn:lb2} is differently weighted. The weight factor is tuned considering the overall reconstruction efficiency of VAE.

%Combining the VAE with the RNN also requires a modification in the objective function to correctly model sequences [REF, STORN]. (recurrent version of the VAE VRNN)  (timestep-wise variational lower bound). In our implementation, however, we simply fine tune between the reconstruction term  by introducing a weight factor for the KL divergence term. The weight is determined by. There are very good arguments why the VAE is chosen over AE. Use of VAE is justified. [REF:ERHAN]

\subsubsection{Reliability Model}
After obtaining the non-linear embedding from the VAE, we train a model (\emph{reliability model}) with the RNN architecture in Eq.~\ref{eqn:rnn_model} using $(\vec{z}^{(i)}, y^{i})$ where $\vec{z}^{(i)}=f_{embed}(\vec{x}^{i})$ and $\vec{x}^{i} \in  \mathcal{D}_{L}$. The resulting reliability model is used to predict RUL for a given set of historical CM data. It is noted the resulting model is similar to the \emph{latent feature discriminative model} (M1), which shows the state-of-the-art performance in various classification tasks~\cite{dgm2014}. The biggest difference lies in the way the non-linear embedding is obtained. The M1 model uses approximate samples from the posterior distribution as an embedding. In our case we rather use the $k^{th}$ layer output as an embedding as we want to avoid using the approximate sampling in order to achieve high precision RUL prediction\footnote{We tried with the approximate samples from the posterior distribution as features to train the reliability model and found the prediction precision is lowered because of the stochastic nature of sampling.}. The corresponding procedure is illustrated in the architectures of the embedding network and the reliability model in Fig.~\ref{fig:embedding}(a) and \ref{fig:embedding}(b).
%% Entire procedure %% Illurstration!
% Phase 1: Generative modeling
% Phase 2: Embedding
% Phase 3:

The entire procedure of our approach from generative modeling to supervised learning of reliability model is given in Algorithm~\ref{sslalgo}, where the mini-batch $\vec{x}^{\textit{M}} = \{\vec{x}^{(i)}\}_{i=1}^{\textit{M}}$ is a randomly drawn sample of size $M$. The $\mathcal{L}$ indicates a corresponding loss function with $\widetilde{\mathcal{L}}$ implying that it is calculated for the data points in mini-batch as an approximation to the quantity for the entire data points. For the VAE, for example,
\begin{equation}\label{eqn:mbloss}
\mathcal{L}(\theta,\phi; \vec{x})  \simeq \widetilde{\mathcal{L}}(\theta,\phi; \vec{x}^{M}) = \frac{n}{M}\sum_{i=1}^{n} \mathcal{L}(\theta,\phi; \vec{x}^{(i)})
\end{equation}
%\simequ \nabla_{\theta,\phi} \widetilde{\mathcal{L}}^{M}_{VAE}
%As explained heretofore, the procedure is comprised of three steps. The first step is the unsupervised training of VAE. The second step is the non-linear embedding. Finally, the last step is a supervised training for reliability model.

\subsubsection{Self-learning Approach}
%As mentioned in Section~\ref{intro}, self-learning also known as self-labeling is a heuristic semi-supervised learning approach.
Self-learning also known as self-labeling is an iterative algorithm that uses supervised learning in repetition~\cite{ssl2010}. As the name suggest, the algorithm tries to ``bootstrap'' using its own prediction. We use the simplest version of self-learning algorithm in order to see if it can help the generalization of supervised learning for the reliability prediction. It is done by first building a reliability model by training a RNN model using the labeled data, $(\vec{x}^{(i)}, y^{i}) \in  \mathcal{D}_{L}$. The resulting model, so called \emph{base learner}, is then used to make RUL estimations for the un-labeled data $\vec{x}^{(j)} \in \mathcal{D}_{U}$. The estimated RUL, $\widetilde{y}^{(j)}$, is then used in the next iteration of training such that the base learner is trained again with $(\vec{x}^{(i)}, y^{i})$ and $(\vec{x}^{(j)}, \widetilde{y}^{(j)})$, treating $\widetilde{y}^{(j)}$ as real labels. The iteration is repeated until a certain convergence condition is met. In our study, we use a single iteration since we do not impose any convergence condition for the algorithm.

\begin{figure*}[t!]
\begin{center}
	\includegraphics[width=0.99\linewidth]{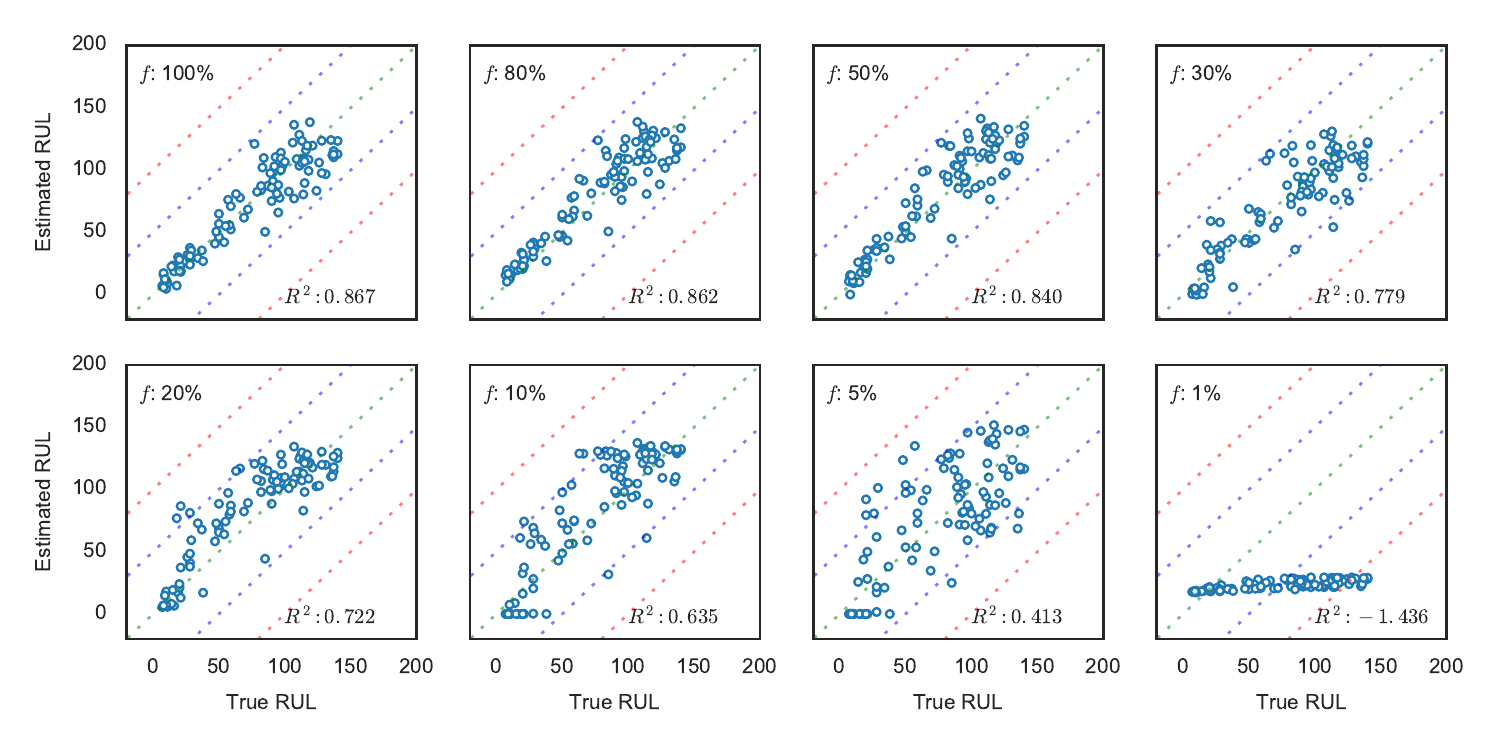} \\ \vspace{-6mm}
	\caption{The actual and the estimated RULs in different labeled fraction scenarios. $f$ is the labeled fraction. The green-dotted line indicates perfect estimation where the difference between the actual and the estimated RULs is zero. The blue and red dotted lines indicate the difference of 30 and 50 cycles, respectively. $R^{2}$ score is also shown.}
	\label{fig:rul_dist}
\end{center}
\vspace{-3mm}
\end{figure*}

\let\oldnl\nl% Store \nl in \oldnl
\newcommand{\nonl}{\renewcommand{\nl}{\let\nl\oldnl}}

\begin{algorithm}
\caption{Semi-supervised learning based on non-linear embedding with VAE. We use settings $M$, $N$ = 100 and $k$ = 3. The first phase follows the minibatch version of the AEVB algorithm~\cite{vae2013} }\label{sslalgo}
\SetKwProg{Func}{Phase 1}{}{end}
\nonl \Func{Generative Modeling}{
%\KwIn{$\mathbf{x^{(i)}} \in \mathbb{\mathcal{D}}$ }
\KwIn{$\vec{x}^{\textit{M}} = \{\vec{x}^{(i)}\}_{i=1}^{\textit{M}} \in \mathbb{\mathcal{D}}$}
% is a randomly drawn sample of $M$. %$\mat{x}=\{\mathbf{x^{(1)}},\cdots,  \mathbf{x^{(n)}} \}$
%\mathbf{x^{(i)}} \in \mathbb{\mathcal{D}}$ }
$\theta$, $\phi$ $\leftarrow$  Initialize parameters\;
\Repeat{convergence of parameters ($\theta$,$\phi$)}{
      $\vec{x^{\textit{M}}}\leftarrow$ Random mini-batch of size, $M$ \;
      $\vec{z^{\textit{M}}}  \sim q_{\phi}(\vec{z^{\textit{M}}}|\vec{x^{\textit{M}}})$ \;
      $\vec{g} \leftarrow \nabla_{\theta,\phi} \widetilde{\mathcal{L}}^{M}_{VAE}$\;
      $\theta$, $\phi$ $\leftarrow$ Update parameters using gradient $\vec{g}$ (e.g., Stochastic Gradient  Descent) \;
    }
\Return $\theta$, $\phi$\;
}
\vspace*{.2cm}
%\SetKwProg{Func}{Phase 2}{}{end}
%\nonl \Func{Embedding}{
%\KwIn{$\vec{x}^{\textit{M}} = \{\vec{x}^{(i)}\}_{i=1}^{\textit{M}} \in \mathbb{\mathcal{D}}_{L}$}
%\Return $\vec{z^{\textit{M}}_{k}} \leftarrow f_{embed, k}(\vec{x^{\textit{M}}}; \theta,\phi)$\;
%\Repeat{all $\vec{x}$ is embedded}{$\vec{z^{\textit{M}}_{k}} \leftarrow f_{embed, k}(\vec{x^{\textit{M}}}; \theta,\phi)$}
%\Return $\vec{z}_{k}$ \;
%}
\vspace*{.2cm}
\SetKwProg{Func}{Phase 2}{}{end}
\nonl \Func{Supervised Learning with Embedding}{
%\KwIn{$(\mathbf{x^{\textit{M}}}, y^{\textit{M}}) \in \mathbb{\mathcal{D}}_{L}$ and $\theta$, $\phi$}
\KwIn{$(\mathbf{x^{\textit{N}}}, y^{\textit{N}}) \in \mathbb{\mathcal{D}}_{L}$ and $\theta$}
$\gamma$ $\leftarrow$  Initialize parameters\;
Embedding $\vec{x}^{\textit{N}}$ into the latent space defined by the $k^{th}$ output layer of the encoder: $\vec{z^{\textit{N}}_{k}} \leftarrow f_{embed}(\vec{x^{\textit{N}}}; \theta)$ \;
\Repeat{convergence of parameters ($\gamma$)}{
      $\vec{z^{\textit{N}}}\leftarrow$ Random mini-batch of size, $N$ \;
      $y^{\textit{N}}  \sim f_{RNN}(\vec{x^{\textit{N}}}; \gamma)$ \;
      $\vec{g} \leftarrow \nabla_{\gamma}\widetilde{\mathcal{L}}^{N}_{RNN}$\;
      $\gamma$ $\leftarrow$ Update parameters using gradient $\vec{g}$ (e.g., Stochastic Gradient  Descent)
    }
\Return $\gamma$\;
}
%\caption{\methodac (\methodfull)}
\label{alg:coltsac}
\end{algorithm}

\section{Experimental Results}\label{exp}
This section presents the evaluation of our approach on a publicly available dataset, \emph{Turbofan Engine Dataset}, provided by Prognostics Center of Excellence in National Aeronautics and Space Administration (NASA)~\cite{saxena2008turbofan}. The dataset is composed of simulated multi-sensor readings from a fleet of aircraft engines of the same type, created with a realistic damage propagation modeling based on C-MAPSS~\cite{saxena2008damage}. The dataset is further divided into training and testing sets. We use the training data to build models for RUL estimation and use the testing data to assess the performance of the models with different metrics. For the evaluations of the proposed approach, we consider the case where the true historical RUL is partially known in the training set. It is done by randomly dropping RUL information for a fraction of the engines in the training samples. We vary the fraction in order to cover a multitude of conditions.

%$By varying the fraction, we test our approach in different conditions.

%By increasing we test out approach in increasingly more challenging situation By varying the labeled fraction from 100\% down to 1\%, we test our approaches in increasingly harsher conditions.

%While we use the training data to build models for RUL estimation and use the testing data to assess the performance of the models with different metrics, we consider the case where the RUL is partially known in order to study semi-supervised approaches.

%The training data consists of the engines with are used to train a model for RUL estimation, which is used to infer RUL for the units in testing set whose neither initial wear nor current wear is not know.
%As mentioned earlier, we consider the case where the RUL is partially known in order to study semi-supervised approaches. It is done by randomly dropping RUL information for a fraction of the engines in training samples.
%Objective %obtained with damage propagation modeling  from run-to-failure simulation  datasets, C-MAPSS Turbofan Engine Dataset. The datasets are composed of simulated multi-sensor readings from a fleet of aircraft engines of the same type, obtained from the run-to-failure simulation of degradation (with damage propagation modeling).

%\begin{figure}
%\begin{center}
%	\includegraphics[width=0.99\linewidth]{FIG/perf_as_frac.pdf} \\ \vspace{-3mm}
%	\caption{Pearson correlation coefficient $\rho$}
%	\label{fig:metrics}
%\end{center}
%\end{figure}

%, i.e., the variability of means

\subsection{C-MAPSS Turbofan Engine Dataset}

The dataset, \textit{FD001}, consists of multiple multivariate (24 sensors including 3 operational settings) time-series from a fleet of engines with different degree of initial wear and manufacturing variation that are unknown. The training and the testing sets contain 100 such engines for each. For the engines in the training set, run-to-failure time-series trajectories are provided. Whereas for the engines in the testing set, time-series trajectories are truncated prior to failures with true RUL values at the point of truncation given in a separate file for validation. A total number of cycles in the training set and in the testing set reach about 20K and 13K, respectively.

\subsection{Evaluations}

% 전체적인 Evaluation 방법에 대해서 out-line할 것!
\subsubsection{Overview}
We directly model the engine failure by parameterizing the RNN architecture in Eq.~\ref{eqn:rnn_model} such that the input sensor readings are directly mapped to historical RUL. RUL is assigned to each sequence for a given engine by simply counting the number of cycles left before failure (by definition). In this way, we avoid to create any domain-specific assumptions for the modeling. We fist build a reliability model in a purely supervised setting, where all the training samples are used with their corresponding RULs. We then omit the RULs for a randomly chosen set of engines in the training set and repeat the procedure.
%We then omit the labels for a fraction of training samples (e.g., omitting all RUL for 30 randomly chosen turbofan engines) and repeat the training.
The faction of engines, for which the RUL information is dropped, is varied from 100\% down to 1\% ($f$ = 100, 80, 50, 30, 20, 10, 5, and 1\%). To remove any selection bias, we repeat the random selection five times for each fraction. The performance measures are then averaged over the five different sets of samples except for the 100\% labeled scenario, where no selection is involved. Also in order to take the model variability into account, we always use the homogeneous ensemble of the models obtained from five independent training, which in general produce a better result.

%We choose the labeled fractions to be 1, 5, 10, 20, 50, 80, 90, and 100 \% of the entire engines.

%In this way, we avoid to make additional assumptions, which to some extent domain-specific. to minimize domain-specific assumptions. %learning a mapping input variables to a target output of RUL. W %The degradation is modeled from the run-to-failure samples (i.e., training), from which a RUL is inferred for a specific point in time. Instead of modeling
%we model RUL directly without any intermediate step to model degradation via health index .

\subsubsection{Preprocessing}

%Minimum possible assumption is made, .. "to estimate RUL without any prior knowledge of underlying physics the system"

Based on the 3 operational settings, we identify 14 different operational modes that are uniquely defined. We normalize the 21 sensor readings (excluding the 3 operational settings) before feeding into our models to train. The normalization is done for each sensor based on min-max scaling and it is done operational-mode-wise such that each sensor is scaled to have a range between 0 and 1 in each operational mode. The sensors that are constant (i.e., $\sigma_{x}=$ 0) after the normalization are dropped. After checking carefully with its impact on the prediction accuracy, we further drop \emph{discrete} sensors with the number of unique values falling below 20. For the sake of simplicity, the discrete variables were excluded in the training of VAE. For the RUL prediction, we limit the value below a maximum RUL of 140 cycles as suggested in~\cite{lstmed, heimes2008recurrent}. Consequently, we limit the assigned RULs in the training set by setting those that exceed the maximum RUL to be the maximum.

%This procedure is done to simply the VAE training where continuous real valued inputs are assumed.  %so that the variational autoencoder can model continuous values.  By carefully checking the performance
%\begin{align}
%\begin{aligned}\label{eq:norm}
%x_{norm}^{(d,m)} = \frac{x^{(d,m)} - x_{min}^{(d,m)}}{x_{max}^{(d,m)} - %x_{min}^{(d,m)}},
%\end{aligned}
%\end{align}

%\subsubsection{Model tuning}

\begin{figure*}[t!]
\begin{center}
\includegraphics[width=0.99\linewidth]{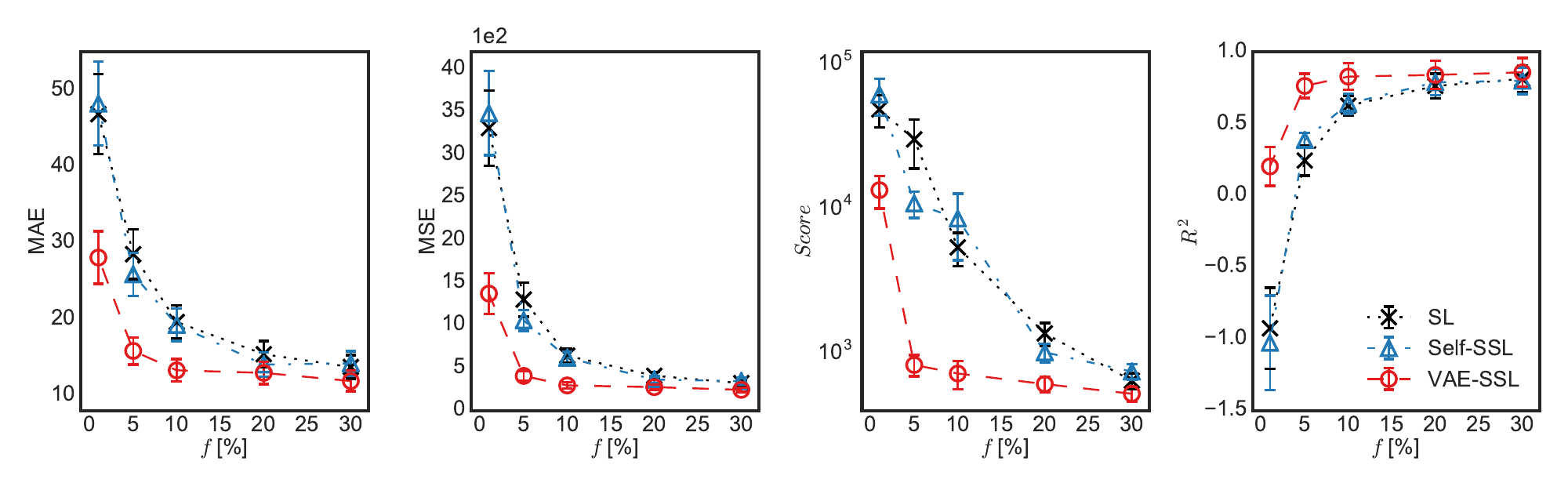} \\ \vspace{-5.5mm}
\subfloat[\label{mae} $MAE$]{\hspace{.25\linewidth}}
\subfloat[\label{mse} $MSE$]{\hspace{.25\linewidth}}
\subfloat[\label{score} $Score$]{\hspace{.25\linewidth}}
\subfloat[\label{r2} $R^{2}$]{\hspace{.25\linewidth}}
	\caption{Measured performance of different approaches as a function of the labeled fraction: (a) $MAE$, (b) $MSE$, (c) $Score$, and (d) $R^{2}$. The error bar indicates the standard error of mean at each point.}
	\label{fig:finalresult}
\end{center}
\vspace{-3mm}
\end{figure*}

\subsubsection{Metrics}
Different performance metrics are employed to assess the performance of the models. This includes the mean squared error ($MSE$), mean absolute error ($MAE$), and the score metric proposed in~\cite{saxena2008damage}. The proposed score metric is defined such that late predictions are more heavily penalized than early prediction:
\begin{equation}\label{eqn:score}
S = \sum_{i=1}^{n}(\exp{(\alpha|\Delta_{i}|)}-1)
\end{equation}
%\begin{align}
%\begin{aligned}\label{eq:score}
%S = \sum_{i=1}^{n}(\exp{(\alpha|\Delta_{i}|)}-1)
%\end{aligned}
%\end{align}
where $\Delta_{i}$ is the difference between the estimated and the true RUL values, i.e., estimated RUL - true RUL, for the $i^{th}$ engine in the testing set and $\alpha$ is equal to $1/13$ if $\Delta_{i} < 0$ or $1/10$ otherwise. We also use the coefficient of determination also known as $R^{2}$ score as a different measure of regression quality\footnote{$R^{2}$ can yield negative values as models can be arbitrary worse.}.

%The proposed score metric is defined as below such that late predictions are more heavily penalized than early prediction

%% R2 metric / defintion
%The asymmetric scoring function, which is defined as XX, reflects . An early prediction is preferred over late prediction as stated in [MAPSS]/ Asymmetric.

%\label{tab:compare_methods}
%\caption{Comparison of \methodac and other algorithms. Our main proposed method \methodac shows the best performance for all metrics. For Corr.~Coef.~and Error, we compare at the same memory usage, and for the memory usage, we do at the same accuracy.

%$SL$ indicates supervised learning only.
%$Self-SSL$ is the self-learning technique. $VAE-SSL$ is the proposed technique.

\subsection{Results}

\subsubsection{Reconstruction with Variational Autoencoder}

% VRAE:
% VRNN: https://arxiv.org/pdf/1506.02216.pdf
%As explained in Section XX, VAE is trained such that the input distributions are en.

We use a three-layer architecture in the encoder (17-8-4) and in the decoder (4-8-17) with batch normalization applied to each layer except for the last layer. The architecture is chosen considering the reconstruction error and the performance of the non-linear embedding. The use of batch normalization is considered essential for training VAE with when the architecture becomes deep (i.e., $L > 2$)~\cite{vaetrain2016}. As mentioned earlier, we incorporate the RNN architecture into the VAE such that latent variables are modeled with the dynamic hidden state of a RNN architecture. This requires a modification in the objective function in Eq~\ref{eqn:lb2} so that sum of loss is averaged in temporal dimension. The model is trained end-to-end using the Adam optimizer~\cite{adam2014} with a mini-batch size of 100.

%In order to account for the temporal dependency of the sensor reading, we use recurrent layers in the VAE architecture such that latent variables are modeled with the dynamic hidden state of a RNN architecture. %The resulting VAE is similar to the proposed model in [REF1, REF2].

Fig.~\ref{fig:cycles} shows four sample sensor readings for several engines in sequence overlaid with the reconstructed sensor readings from VAE. In Fig.~\ref{fig:freq}, we show the frequency of the values in histograms to see distributions. The results clearly demonstrate even with the one-fourth of the original input dimension in the bottleneck layer of VAE, the reconstruction is done reasonably well, capturing time dependent features of the signals. The sum of reconstruction loss and the KL-divergence stays below 0.3 when the trained VAE is used to reconstruct the validation sets\footnote{A fraction of the training data is kept from training and used for validation}. This may support the manifold hypothesis that high dimensional data tend to lie in the vicinity of a low dimensional manifold~\cite{dlbook2016}.

\subsubsection{Baseline Results}

% Table  |   MLP  | GRU | ConvLSTM
% MAE
% MSE
% Score

% Performance as a function of labeled fraction
% Details about the baseline models (architecture/epochs/batch/optimization)
% Reconstruction efficiency (MSE?)
% Sample plots (Truth and Reconstructed)
% Histogram / Scattered as a time series for several representative sensors

Fig.~\ref{fig:rul_dist} shows the baseline results of RNN-based models with a four-layer architecture\footnote{We adopt both GRU and LSTM architecture in the models. However, as we find no quantitative difference in performance, we choose the simpler architecture, namely, GRU for the final results.} trained in a purely supervised setting. It shows the true RUL in the $x$-axis and the estimated RUL in the $y$-axis in different labeled fraction scenarios. From the top left to the bottom right, the labeled fraction, $f$, is varied from 100\% to 1\%. When the full training set is used, the trained model shows good performance with the measured $MSE$, $MAE$, $Score$ and $R^{2}$ of 228, 11.3, 345, and 0.87, respectively. This result is comparable to that of other approaches~\cite{lstmed, babu2016deep, khelif2014rul, peng2012modified, ramasso2014investigating}.

Now as the fraction changes, the prediction accuracy decreases. It can clearly be seen from Fig.~\ref{fig:rul_dist} that the correlation between the actual and the estimated RULs get worse and worse with decreasing fraction, also indicated by the $R^{2}$ score shown in the figure. The performance degradation is expected because the complexity of model requires a large amount of training data. In the current implementation of the RNN-based model, the number of tunable parameters reaches few hundreds. Nevertheless, the performance degradation is not prominent until the fraction is decreased down to 30\%, indicating that only 30\% of the training samples are already sufficient for the model to give a reasonably good result. As such, we rather focus in the region below 30\%, where the performance degradation is significantly showing.

%When , i.e., $f$ = 100\%, the model shows good performance with the MSE, MAE, and score of XX, YY, and ZZ, which is comparable to the results with different approaches [see Appendix A in LED]. We note the measured MSE implies the RUL estimation is only off by XX cycles on average.
%It can clearly be seen the model (prediction) accuracy assessed in various metrics degrades as the number of available labels decreases.  the true and the estimated RULs in different labeled fraction scenarios.In order to evaluate our approach, we first build a model in a purely supervised setting using GRU (or LSTM) as explained in Section XX. Fig XX shows the actual and the estimated RULs in different labeled fraction scenarios.

\begin{center}
\begin{table*}[t]
%\begin{tabular}{SSSSSSSSS} \toprule
\caption{Comparison of different approaches. The best result is highlighted in bold.}\label{tab:results}
\begin{tabular}{ c c c c c c c c} \toprule
    {Metric} \hspace{0.5cm} & {Method} & 1\% & 5\% & 10\% & 20\% & 30\% & 100\%\\ \midrule
      		& SL  & 46.9 $\pm$ 5.3 & 28.5 $\pm$ 3.3 & 19.7 $\pm$ 2.2 & 15.4 $\pm$ 1.7 & 13.7 $\pm$ 1.5 & \textbf{11.3} $\pm$ 1.3 \\
    \textit{MAE} \hspace{0.5cm} & Self-SSL  & 48.2	 $\pm$ 5.5  & 25.8 $\pm$ 2.8 & 19.2 $\pm$ 2.1  & 14.0 $\pm$ 1.6  & 14.2 $\pm$ 1.6  & \textbf{11.3} $\pm$ 1.3\\
      		& VAE-SSL  & \textbf{28.1} $\pm$ 3.4 & \textbf{15.8} $\pm$ 1.8 & \textbf{13.3} $\pm$ 1.5 & \textbf{12.9} $\pm$ 1.5  & \textbf{11.9} $\pm$ 1.4 & \textbf{11.3} $\pm$ 1.3\\ \midrule
 %%%%%%%%%%
    		& SL  & 3309 $\pm$ 441 & 1299 $\pm$ 199 & 642 $\pm$ 80   & 407 $\pm$ 45 & 344 $\pm$ 35 & 228 $\pm$ 27\\
    \textit{MSE} \hspace{0.5cm} & Self-SSL  & 3483 $\pm$ 493  & 1055 $\pm$ 120 & 618 $\pm$ 74  & 362 $\pm$ 40   & 344 $\pm$ 38 & 228 $\pm$ 27 \\
      		& VAE-SSL   & \textbf{1370} $\pm$ 240 & \textbf{405} $\pm$ 59 & \textbf{294} $\pm$ 36 & \textbf{274} $\pm$ 31 & \textbf{243} $\pm$ 27 & \textbf{221} $\pm$ 26  \\ \midrule
 %%%%%%%%%%   				
    		%& SL  & 48163.299$\pm$12071.701   & 29937.252$\pm$11145.473 & 5362.855$\pm$1392.765  & 1363.948$\pm$243.205 & 642.528$\pm$80.819 & 0$\pm$0\\
			& SL  & 482 $\pm$ 121   & 299 $\pm$ 111 & 53.6 $\pm$ 13.9  & 13.6 $\pm$ 2.43 & 6.42 $\pm$ 0.81 &\textbf{3.45} $\pm$ 0.41\\    		
    \textit{Score} [$\times$\num{e-2}] \hspace{0.5cm} & Self-SSL  & 610 $\pm$ 173  & 109 $\pm$ 22.4 & 84.9 $\pm$ 41.3   & 10.0 $\pm$ 1.46 & 7.42 $\pm$ 0.93 & \textbf{3.45} $\pm$ 0.41 \\
    & VAE-SSL  & \textbf{133} $\pm$ 34  & \textbf{8.26} $\pm$ 1.38 & \textbf{7.20} $\pm$ 1.58  & \textbf{6.09} $\pm$ 0.72 & \textbf{5.22} $\pm$ 0.59 & 4.75 $\pm$ 0.56  \\ \midrule
   		%& VAE-SSL  & 13335.487$\pm$3364.378  & 825.803$\pm$137.646 & 720.559$\pm$158.243  & 608.768$\pm$72.095  & 522.360$\pm$59.215  & 0.00$\pm$0.00  \\ \midrule   		
 %%%%%%%%%%	
    		& SL  & $-$0.93 $\pm$ 0.28  & 0.24 $\pm$ 0.11 & 0.63 $\pm$ 0.07  & 0.76 $\pm$ 0.09 & 0.81 $\pm$ 0.09 & \textbf{0.87} $\pm$ 0.10 \\
    $R^{2} $\hspace{0.5cm} & Self-SSL  & $-$1.03 $\pm$ 0.33  & 0.39 $\pm$ 0.05 & 0.64 $\pm$ 0.07  & 0.79 $\pm$ 0.09  & 0.80 $\pm$ 0.09  & \textbf{0.87} $\pm$ 0.10\\
    		& VAE-SSL & \textbf{0.20} $\pm$ 0.14  & \textbf{0.76} $\pm$ 0.08 & \textbf{0.83} $\pm$ 0.09  & \textbf{0.84} $\pm$ 0.10   & \textbf{0.86} $\pm$ 0.10 & \textbf{0.87} $\pm$ 0.10 \\ \bottomrule
\end{tabular}
\end{table*}
\end{center}

\subsubsection{Semi-supervised Approach}

%that were generated based on a realistic modeling of HPC (?) degradation. The degradation of the a.. modeled by training on samples of The accuracy (performance) of RUL prediction/modeling is assessed by with different performance metrics. Mean Square Error (MSE), Mean Absolute Error (MAE), From the training samples, RUL modeling .. estimated and evaluated with several performance metrics. The data were  that degrade multi-sensor readings  a set of simulated run-to-failure  of aircraft turbofan engine , based on a realistic modeling of .. with different faults modes.  simulation/ degradation of  The details of the dataset shall follow in the next section. We first give the evaluation of baseline models  We then define .. by omitting

As discussed in Section~\ref{ourapproach}, we apply semi-supervised learning techniques to see if the performance can be improved when only a small fraction of the training data is labeled. We try two different semi-supervised learning techniques, i.e., self-learning and VAE-based non-linear embedding. Fig.~\ref{fig:finalresult} shows the results of the RUL estimation quantified by the three different metrics plotted as a function of $f$. From left to right, $MAE$, $MSE$, $Score$, and $R^{2}$, are plotted with corresponding statistical uncertainties. The score is plotted in logarithmic scale.
%envelope the entire range while showing the difference.
It quantitatively shows the performance degradation becomes increasingly larger as the fraction decreases. The result from the self-learning plotted in a triangular shape reveals that the performance is essentially the same within statistical uncertainty except for the score at $f$ = 5\%, where a distinctive improvement is shown.

Now, the result from the non-linear embedding plotted in a circle shows improvements for $f$ below 30. The level of improvement becomes increasingly larger reaching x1.7, x2.4, and x3.6 improvement, respectively for $MAE$, $MSE$, and $Score$. The $R^{2}$ improvement is also notable. We note the prediction accuracy of the model is kept rather high even down to 5\% fraction. The numerical compilations of the measured performance metrics are given in Table~\ref{tab:results} for all three approaches: $SL$ indicates supervised learning only. $Self$-$SSL$ is the self-learning. $VAE$-$SSL$ is the proposed method.

% even with 1\% fraction the model prediction accuracy is rather high %; the average difference between the actual and the estimated is about XX.  %To make sure the observed improvements are not the artifacts of , we vary related parameters % This implies when basic learner , i.e., model bootstrapping is not. %Score blows up, it is interesting to point out that how dangerous it could be  a complicated DNN model not regularized.

%\begin{center}
%\label{tab:results}
%\caption{Results}
%\begin{tabular}{ |c|c|c|c| }
%\hline
%col1 & col2 & col3 \\
%\hline
%\multirow{3}{4em}{Multiple row} & cell2 & cell3 \\
%& cell5 & cell6 \\
%& cell8 & cell9 \\
%\hline
%\end{tabular}
%\end{center}

%\begin{table}
%\begin{center}
%\label{tab:results}
%\caption{Results}
%\scriptsize
%\begin{tabular}{lcccc} \toprule[0.1em]
%	\bottomrule[0.1em]
%\end{tabular}
%\end{center}
%\end{table}

\section{Discussion}
% Conclusions and Future works

% The contribution of the paper is twofold /our main contribution is twofold
% First,
% Second,
% address the issue of  also known as No Fault Found (NFF) problem [REF].

This paper described a semi-supervised learning approach to predict asset failures when there exist relatively small label information available. Our approach uses the non-linear embedding based on a deep generative model. We used the variational autoencoder as a deep generative model choice because of its relatively easy trainability on top of its firm theoretical foundation. The VAE was trained unsupervised while utilizing all the available data whether it is labeled or unlabeled. With this approach, we pursued achieving good prediction accuracy even when the available label information is highly limited.

%Our approach is based on non-linear embedding obtained from a deep generative model.
%% How to make use of unlabled data in building a complicated deep learning model with supervised training. %% why is this important? we argue / our observation. / maybe later %% evaluation / we argue  %% Our obeservation suggested while such situation is common (commonly encountered), there have been relatively little attention paid to.

The approach was extensively evaluated on a publicly available Turbofan Engine dataset by varying the fraction of labels down to 1\%. The evaluation shows significant improvements in prediction accuracy when compared to the supervised and a naive semi-supervised approach, i.e., self-learning at 1\%, the improvement reaches up to x3.6 for the proposed score metric. While the 1\% fraction could be considered rather extreme, it is not unusual in practice since the acquisition of exact health status information that can be related to failures is exceedingly difficult in many cases. In image recognition tasks, order of 0.1 to 1\% labeled fraction is typically considered for similar studies.

%% NFF case as well.
Overall, our study suggests the semi-supervised learning approach with the non-linear embedding based on deep generative model is effective when dealing with the problem of insufficient labels in future reliability prediction. Such problem is of tremendous practical interest in a wide range of data-driven PHM applications, especially when sophisticated deep learning architectures are used in modeling. In principle, there is little reason why our approach can not be extended to use other deep generative models for the embedding. In ensuing efforts, we are extending the approach to incorporate non-linear embedding using GAN.

\vspace{-2mm}
{ %\small
\fontsize{7.3}{8.76}\selectfont
\bibliographystyle{abbrv}
\bibliography{kdd_skt.bib}
}

%\appendix
%\input{099appendix}
\end{document}